\documentclass[runningheads]{llncs}
\usepackage{float}
\usepackage{graphicx}
\usepackage{amsmath}
\usepackage{amsfonts}
\usepackage{xcolor}
\usepackage{url}
\usepackage{multirow}
\usepackage{array}
\newcolumntype{C}[1]{>{\centering\arraybackslash}p{#1}}
\usepackage{grffile}
\usepackage{rotating}

\usepackage{cite}

\usepackage{booktabs}

\begin{document}

\title{Best Practices for a Handwritten Text Recognition System}

\author{George Retsinas\inst{1} \and Giorgos Sfikas\inst{2, 3, 4} \and Basilis Gatos\inst{2} \and Christophoros Nikou\inst{3}}
\authorrunning{Retsinas et al.}

\institute{
School of Electrical and Computer Engineering \\
National Technical University of Athens, Greece \\
\and
Computational Intelligence Laboratory \\ 
Institute of Informatics and Telecommunications, National Center for Scientific Research ``Demokritos", Greece \\
\and
Department of Computer Science and Engineering \\
University of Ioannina, Greece \\
\and
Department of Surveying and Geoinformatics Engineering \\
University of West Attica, Greece \\
\email{gsfikas@uniwa.gr, gretsinas@central.ntua.gr, cnikou@cse.uoi.gr, bgat@iit.demokritos.gr}\\
}

\maketitle
\begin{abstract}
Handwritten text recognition has been developed rapidly in the recent years, following the rise of deep learning and its applications. 
Though deep learning methods provide notable boost in performance concerning text recognition, non-trivial deviation in performance can be detected even when small pre-processing or architectural/optimization elements are changed. 
This work follows a ``best practice'' rationale; highlight simple yet effective empirical practices that can further help training and provide well-performing handwritten text recognition systems. 
Specifically, we considered three basic aspects of a deep HTR system and we proposed simple yet effective solutions: 
1) retain the aspect ratio of the images in the preprocessing step, 
2) use max-pooling for converting the 3D feature map of CNN output into a sequence of features and 
3) assist the training procedure via an additional CTC loss which acts as a shortcut on the max-pooled sequential features.
Using these proposed simple modifications, one can attain close to state-of-the-art results, while considering a basic convolutional-recurrent (CNN+LSTM) architecture, for both IAM and RIMES datasets. 
Code is available at \url{https://github.com/georgeretsi/HTR-best-practices/}.

\end{abstract}

\section{Introduction}
\label{sec:introduction}

Handwritten Text Recognition (HTR) is an active area of research, combining ideas from both computer vision and natural language processing.  
Unlike recognition of machine-printed text, handwriting is related to a number of unique characteristics that make the task much more challenging than traditional optical character recognition (OCR).
The challenging nature of handwriting recognition stems mostly from the potentially high writing variability between individuals.
To this end, along with visually decoding an image into sequence of characters, several HTR works adopt language models to reduce this innate ambiguity of handwritten characters, making use of contextual and semantic information.

In general, designing an effective and generalizable learning system is a ongoing challenge, 
with transferability between different learned writing styles more being not a given in most cases \cite{retsinas2018transferable}.
Neural Networks (NNs), among a variety of other learning systems, have been used for the recognition of handwriting from early on, with a span ranging between simpler subtasks such as single digit recognition \cite{bishop2006pattern} up to full, unconstrained offline HTR~\cite{fischer2012handwriting, puigcerver2017multidimensional}. 
Following the rise of deep learning and its applications, recent developments in HTR are monopolized by Deep Neural Networks (DNNs). 
The seminal work of Graves et al.~\cite{graves2006connectionist} played a pivotal role in the rise of deep learning for HTR applications by enabling the training of recurrent nets without assuming any prior character segmentation. 
A plethora of subsequent works for HTR relied on Graves et al. in order to train modern and notably effective DNNs~\cite{puigcerver2017multidimensional, krishnan2018word, retsinas2021deformation, luo2020learn}.

This work focuses on finding best practices for building modern HTR systems. 
We explore a set of guidelines for training HTR DNNs, re-examining and extending ideas from several previous works of ours \cite{retsinas2021seq2seq,retsinas2021iterative,retsinas2021deformation}.
We start with a fairly common deep network architecture for HTR, consisting of a CNN backbone and a BiLSTM head, 
and we make simple yet effective architectural and training choices. 
These best practice suggestions can be categorized and summarized as follows: 
\begin{enumerate}
    \item \textbf{pre-processing:} retain aspect ratio of images and use batches of padded images in order to effectively use mini-batch Stochastic Gradient Descent (SGD)
    \item \textbf{architectural:} replace the the column-wise concatenation step between the CNN backbone and the recurrent head with a max-pooling step. 
    Such a choice not only reduces the required parameters but has an intuitive motivation: 
    we care only about the existence of a character and not its vertical position.
    \item \textbf{training:} add an extra shortcut branch, consisting of a single 1D convolution layer, at the output of the CNN backbone. 
    This branch results to an extra character sequence estimation, trained in parallel to the recurrent branch. 
    Both branches use a CTC loss.
    The motivation behind such a choice comes from the increased difficulty of training recurrent layers. 
    However, if such a straightforward shortcut exists, the output of the CNN backbone should converge to more discriminative features, 
    ideal for fully harnessing the power of recurrent layers compared to an end-to-end training scheme.  
\end{enumerate}

The contribution of this paper is best highlighted through the experimental section, where we achieve state-of-the-art results with the aforementioned choices, despite the simplicity of the employed network.
Furthermore, other state-of-the-art existing methods propose complex architectures and augmentation schemes which are orthogonal to our approach, highlighting the importance of the suggested best practices.


\section{Related Work}

As is the case with most, if not all, tasks in computer vision, modern HTR literature is dominated by neural network-based methods.
Recurrent neural networks have become the baseline \cite{leifert2016cells, puigcerver2017multidimensional}, as they naturally fit to the sequential nature of handwriting.

Recurrent-based approaches have thus practically overshadowed the previous state-of-the-art, which was based mostly on Hidden Markov Model (HMM)-based approaches.
Since the introduction of the standard recurrent network paradigm \cite{fischer2012handwriting, fischer2012lexicon}, many key advances have emerged paving the way for very efficient HTR systems. 
A characteristic example is the integration of the Long Short-Term Memory models (LSTMs) into HTR systems~\cite{greff2016lstm}, that effectively dealt with the vanishing gradient problem. 
More importantly, Graves et al.~\cite{graves2006connectionist} introduced a very effective algorithm algorithm for training such HTR systems with sequence-based loss using dynamic programming.  
Specifically, this Connectionist Temporal Classification (CTC) method and corresponding output layer \cite{graves2012connectionist}, a differentiable output layer that maps a sequential input into per-time unit softmax outputs, allows simultaneous sequence alignment and recognition with a suitable decoding scheme. 
Multi-dimensional recurrent networks have been considered for HTR \cite{leifert2016cells}, 
however there has been criticism that the extra computational overhead may not translate to an analogous increase in efficiency~\cite{puigcerver2017multidimensional}.

Even though in this work we will focus only on greedy decoding of a CTC-trained network, research on decoding schemes is also active \cite{collobert2019fully}, with the beam search algorithm being a popular approach, capable of exploiting an external lexicon as an implicit language model. 


Sequence-to-Sequence approaches, involving translating an input sequence to an output sequence of a different length in general, became very popular when achieved state-of-the-art results in Natural Language Processing and gradually evolved to Transformer networks with attention mechanisms~\cite{vaswani2017attention}. 
Such approaches were later adopted successfully by the HTR community~\cite{sueiras2018offline, chowdhury2018efficient, michael2019evaluating, wick2021transformer, retsinas2021seq2seq}. 

Recent research directions include complex augmentation schemes (~\cite{krishnan2018word, wigington2017data, luo2020learn}), novel network architectures/modules (e.g. Seq2Seq/Transformers, Spatial Transformer Networks~\cite{dutta2018improving}, deformable convolutions~\cite{retsinas2021deformation}) and multi-task losses with auxiliary training feeds (e.g. n-gram training~\cite{tassopoulou2021enhancing}). 



\section{Proposed HTR System}
\label{sec:proposed}

In what follows, we will describe in detail the proposed HTR system with emphasis given on the suggested best practice modifications. 
The described system takes as input either a word or a line image and then returns the predicted sequence of characters based on an unconstrained greedy CTC decoding algorithm~\cite{graves2012connectionist}.

\subsection{Preprocessing}

The pre-processing steps, applied to every image, are: 
\begin{enumerate}
    \item All images are resized to a resolution of $128\times 1024$ pixels for line images or $64\times 256$ pixels.
Initial images are padded (using the image median value, usually zero) in order to attain the aforementioned fixed size. 
If the initial image is greater than the predefined size, the image is rescaled. 
The padding option aims to preserve the existing aspect ratio of the text images.
\item During training, image augmentation is performed. A simple global affine augmentation is applied at every image, considering only rotation and skew of small magnitude in order to generate valid images. Additionally, gaussian noise is added to the images.
\item 
Each word/line transcription has spaces added before and after, i.e. "He rose from" is changed to " He rose from ". 
This operation aims to assist the system to adapt to the marginal spaces that exist in the majority of the images during the training phase.
For the testing phase, these additional spaces are discarded.
\end{enumerate}

Augmentation operations are part of every modern deep learning system and can consistently provide increased performance, allowing better generalization~\cite{puigcerver2017multidimensional}. 
The used augmentation scheme is very basic, trying to have minimal overhead from this step.

The addition of extra spaces in the transcription is not explicitly referred to recent existing works, but it is intuitive given the pad operation of step 1 which creates large empty margins. It has a minor yet positive impact to our system and thus is added as a step. Due to the reduced significance of this step in the overall performance, it is not explored in the experimental section.

On the other hand, we found the padding operation critical in many settings. 
A typical trade-off, met in many recent text recognition/spotting works, concerns the definition of the input size: using a predefined fixed size can assist the architectural design of CNNs and training time requirements, while retaining the initial image size by processing individually each image (e.g.\cite{sudholt2016phocnet}) may lead to better performance at the cost of discarding the mini-batch option. 

Modern DNN training relies on creating batches of several images, since batch manipulation of images can notably affect the training time by fully utilizing the GPU resources.
Thus image resizing is a widely-used first step for any vision problem when DNNs are involved.
On the contrary, when using different sized images by processing each image individually and update the network's weights after a predefined number of images, as if a batch was processed, leads to an impractical time-consuming training procedure to otherwise lightweight DNNs, where  the existing hardware is under-utilized.

In this work, contrary to the majority of existing approaches, we propose a simple, yet elegant, solution: we aim to retain the aspect ratio of the images and simultaneously organize them into batches.  
The images are transformed into the same, predefined, shape without resizing, if possible. 
Specifically, if the image size is smaller than the predefined size, we pad the image accordingly. 
The padding operation is performed equally at each direction, positioning the initial image at the center of the new one, with a fixed value, the median value of the initial image.
If the image is larger than the predefined size, it is resized,affecting the aspect ratio. 
To assist the proposed approach, we can compute the average height and width over the whole set of the initial images and select an appropriate size in order to perform the aforementioned resize operation scarcely (only for very large words/sentences) and thus avoid deformations that are generated by frequently violating the aspect ratio.

\begin{figure}[htb]
\centering
\includegraphics[width=.95\linewidth]{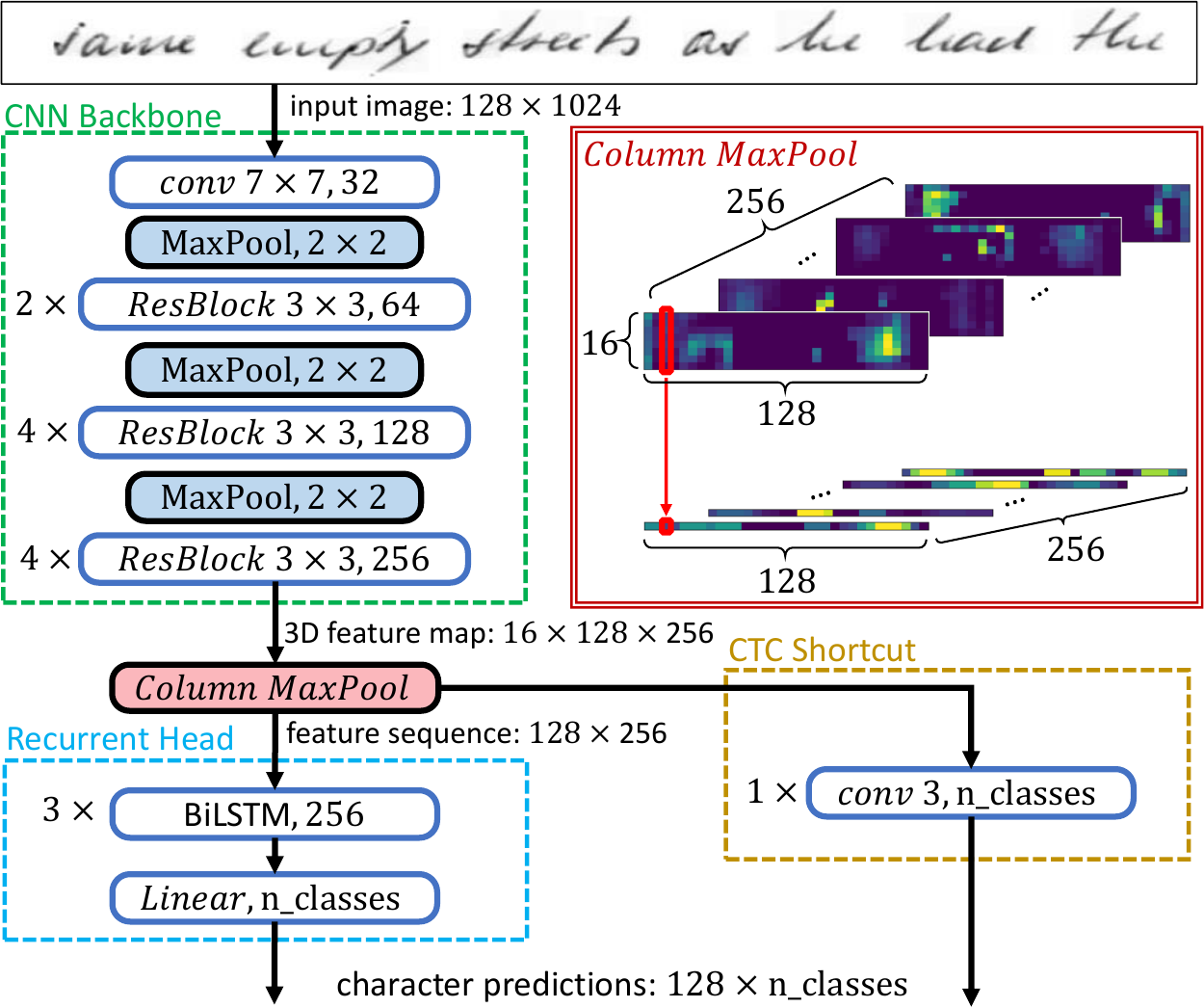} 
\caption{Overview of the DNN architecture. Apart from the CNN backbone, the recurrent head, we also depict the auxiliary CTC shortcut branch which will be the core component of the proposed training modification. Furthermore, we visualize the column-wise max-pooling operation that is performed between the CNN backbone and the recurrent head.}
\label{fig:arch}
\end{figure}

\subsection{Network Architecture}

The model that we will use to test the proposed technique can be characterized as a convolutional-recurrent architecture (an architecture overview is depicted in Figure~\ref{fig:arch}).
The convolutional-recurrent architecture can be broadly defined as a convolutional backbone being followed by a recurrent head, typically connected to a CTC loss.
Convolutional-recurrent variants have given routinely very good results for HTR \cite{dutta2018improving,puigcerver2017multidimensional}.

\subsubsection{Convolutional Backbone:}

In our model, the convolutional backbone is made up of standard convolutional layers and ResNet blocks \cite{he2016deep}, interspersed with max-pooling and dropout layers.
In particular, the first layer is a $7 \times 7$ convolution with $32$ output channels, followed by cascades of $3 \times 3$ ResNet blocks~\cite{he2016deep}: a series of $2$ ResNet blocks with $64$ output channels, $4$ ResNet blocks with $128$ output channels
and $4$ ResNet blocks with $256$ output channels.
The standard convolution and the ResNet blocks are all followed by 
ReLU activations, Batch Normalization and dropout.
Between cascades of blocks we downscale the produced feature map with $2 \times 2$ max-pooling operations of stride $2$, as shown in Figure~\ref{fig:arch}.
Overall, the convolutional backbone accepts a line image and outputs a tensor of size $h \times w \times d$ (e.g. assuming the line image case, the tensor is of size $16 \times 128 \times 256$).

\subsubsection{Flattening Operation:}
The convolutional backbone output should be transformed into a sequence of features in order to processed by recurrent networks. 
Typical HTR approaches, assume a column-wise approach (towards the writing direction) to ideally simulate a character by character processing. 
In our work, the CNN output is flattened by a max-pooling operation in a column-wise manner. 
Flattening of the extracted feature maps by the widely-used concatenation operation would result into a sequence of length $w$ with feature vectors of size $hd$, while max-pooling results to reduced feature vectors of size $d$. 
Apart from the apparent computational advantage, \emph{column-wise max-pooling} achieves model translation invariance in the vertical direction. In fact, the reasoning behind max-pooling is that we care only about the existence of features related to a character and not their spatial position. This has been the major motivation for \emph{column-wise max-pooling}, as successfully employed in our previous works~\cite{tassopoulou2021enhancing, retsinas2021deformation, retsinas2021seq2seq}.

\subsubsection{Recurrent Head:}
The recurrent component consists of $3$ stacked Bidirectional Long Short-Term Memory (BiLSTM) units of hidden size $256$.
These are followed by a linear projection layer, which converts the sequence to a size equal to the number of possible character tokens, $n_{classes}$ (including the blank character, required by CTC). 
The final output of the recurrent part can be translated into a sequence of probability distributions by applying a softmax operation. 
During evaluation, the aforementioned greedy decoding is performed by selecting the character with the highest probability at each step and then removing the blank characters from the resulting sequence~\cite{graves2012connectionist}.

\subsection{Training Scheme}

The training of the HTR system is performed via an Adam~\cite{kingma2014adam} optimizer using an initial learning rate of $0.001$ which gradually decreases using a multistep scheduler. The overall training epochs are 240 and the scheduler decreases the learning rate by a factor of $0.1$ at 120 and 180 epochs. 

This optimizing scheme, with minor modifications, is commonly used for HTR systems. 
Nonetheless, we assume an end-to-end training approach where both the convolutional and the recurrent parts of the system are optimized through the final CTC loss. 
Even though this typical approach produces well-performing solutions, the LSTM head may encumbers the overall training procedure, since recurrent modules are known to exhibit convergence difficulties.

To circumvent this training hindrance, we introduce an auxiliary branch as shown in Figure~\ref{fig:arch}. We dub this extra module as a ``CTC shortcut''. 
In what follows, we describe this module and its functionality in detail. 

\subsubsection{CTC shortcut:}
Architecture-wise, the CTC shortcut module consists only of a single 1D convolutional layer, with kernel size 3. Its output channels equal to the number of the possible character tokens ($n_{classes}$). 
Therefore, the 1D convolutional layer is responsible for straightforwardly encoding context-wise information and providing an alternative decoding path. 
Note that we strive for simplicity for this auxiliary component, since its aim is to assist the training of the main branch and thus a shallow convolutional part of only one layer is ideal for this task. 
We do not expect from the CTC branch to result to precise decodings. 

The CTC shortcut is trained along with the main architecture using a multi-task loss by adding the corresponding CTC losses of the two branches with the appropriate weights. Specifically, if $f_{cnn}$ represents the convolutional backbone, $f_{rec}$ represents the recurrent part and $f_{shortcut}$ represents the proposed shortcut branch, while $I$ is an input image and $s$ its corresponding transcription, the multi-task loss is written as: 
\begin{equation}
    L_{CTC}(f_{rec}(f_{cnn}(I)); s) + 0.1 \, L_{CTC}(f_{shortcut}(f_{cnn}(I)); s)
\end{equation}
Since CTC shortcut acts only as an auxiliary training path, it is weighted by $0.1$ to reduce its relative contribution to the overall loss. 

The motivation behind this extra branch is rather simple: overall convergence is assisted by quickly generating discriminative features at the top of the CNN backbone through the straightforward 1D convolutional path, simplifying the training task for the recurrent part. 
Due to its training-oriented assisting nature, CTC shortcut is used only during training and omitted during evaluation. 
Therefore, this proposed shortcut does not introduces any overhead during inference.

\section{Experimental Evaluation}

Evaluation of the proposed system is performed on two widely used datasets, IAM\cite{marti2002iam} and Rimes \cite{grosicki12008rimes}. 
The ablation study, considering different settings of the proposed methodology, is performed on the challenging IAM dataset, 
consisting of handwritten text from 657 different writers and partitioned into writer-independent train/validation/test sets 
(we use the same set partition as in \cite{puigcerver2017multidimensional}). 
All experiments follow the same setting: line-level or word-level recognition using a lexicon-free unconstrained greedy CTC decoding scheme.
Character Error Rate (CER) and Word Error Rate (WER) metrics are reported in all cases (lower values are better).


\subsection{Ablation Study}

First, we explore the impact of the proposed modifications over both the validation and the test set of IAM dataset. Moreover, both line-level recognition (Table~\ref{tab:line-rec}) and word-level recognition (Table~\ref{tab:word-rec}) are considered. 
Specifically, we investigate the difference in performance when we: 1) use \emph{resized} or \emph{padded} (retain aspect-ratio case) input images, 2) use \emph{concatenation} of \emph{max-pooling} flattening operation between the convolutional backbone and the recurrent head and 3) use or not the \emph{CTC shortcut during} the training process.

\begin{table}[h]
 \caption{Line-level recognition results for IAM dataset: Exploring the impact of the proposed modifications.}
  \centering
 \resizebox{\linewidth}{!}{
  \begin{tabular}{C{.17\linewidth}C{.17\linewidth}C{.17\linewidth}|C{.11\linewidth}C{.11\linewidth}|C{.11\linewidth}C{.11\linewidth}}
    & & & \multicolumn{2}{c}{Validation} & \multicolumn{2}{c}{Test}\\
    \hline
    Preprocessing & Flattening &  CTC Shortcut & CER(\%) & WER(\%) & CER(\%) & WER(\%) \\
    \hline
    \multirow{ 2}{*}{resized} & \multirow{ 2}{*}{concatenation} &   no & 4.28 & 15.29 &	5.93 & 19.57\\
    & &  yes & 3.72 & 13.18 & 5.11  &  16.96\\
    \hline
    \multirow{ 2}{*}{resized} & \multirow{ 2}{*}{max-pooling} & no & 3.73 & 13.54 & 5.28 & 17.77 \\
    &  & yes & 3.47 &	12.77 &	4.85 &	16.19\\
    \hline
    \multirow{2}{*}{padded} & \multirow{ 2}{*}{concatenation} & no & 4.06  & 14.40  & 5.54  & 18.60\\
    & & yes & 3.37 & 12.22 & 4.71 & 15.94\\
    \hline
    \multirow{2}{*}{padded} & \multirow{ 2}{*}{max-pooling} & no & 3.46	& 12.55 &	4.93 & 16.81 \\
    & & yes & \textbf{3.21} & \textbf{11.89}  & \textbf{4.62} & \textbf{15.89} \\
    \hline
  \end{tabular}
  }
  \label{tab:line-rec}
\end{table}

\begin{table}[h]
 \caption{Word-level recognition results for IAM dataset: Exploring the impact of the proposed modifications.}
  \centering
 \resizebox{\linewidth}{!}{
  \begin{tabular}{C{.17\linewidth}C{.17\linewidth}C{.17\linewidth}|C{.11\linewidth}C{.11\linewidth}|C{.11\linewidth}C{.11\linewidth}}
    & & & \multicolumn{2}{c}{Validation} & \multicolumn{2}{c}{Test}\\
    \hline
    Preprocessing & Flattening & CTC Shortcut & CER(\%) & WER(\%) & CER(\%) & WER(\%) \\
    \hline
    \multirow{ 2}{*}{resized} & \multirow{ 2}{*}{concatenation} & no &  4.35 & 12.55 &	5.58 &	15.46\\
    & &  yes & 4.27 &	12.02 &	5.46 &	15.13\\
    \hline
    \multirow{ 2}{*}{resized} & \multirow{ 2}{*}{max-pooling} & no & 4.25	& 12.17 &	5.69 & 15.87\\
    & & yes & 4.09 &	11.65 &	5.23 & 14.40\\
    \hline
    \multirow{ 2}{*}{padded} & \multirow{ 2}{*}{concatenation} & no & 4.17 &	11.99 &	5.66 & 15.66\\
    & & yes & 3.98 & 11.50 &	5.37 & 14.98 \\
    \hline
    \multirow{ 2}{*}{padded} & \multirow{ 2}{*}{max-pooling} & no & 4.00 &	11.25 &	5.43 &	15.06\\
     &  & yes & \textbf{3.76} &	\textbf{10.76} & \textbf{5.14} &	\textbf{14.33}\\
    \hline
  \end{tabular}
  }
  \label{tab:word-rec}
\end{table}

The following observations can be made:
\begin{itemize}
    \item Retaining the aspect-ratio of the images (padded option) achieves improved results for the majority of cases.
    \item Performing the flattening operation via max-pooling not only is more cost-effective, but it has a positive impact on performance. This is more evident in line-level recognition setting.  
    \item Training with a CTC shortcut module provides notable boost over all cases. For example, in line-level recognition the significant difference in performance when considering different flatting operations is considerably decreased when the CTC shortcut approach is adopted (e.g. for padded line-level recognition the WER  performance difference drops from 1.79\% to only 0.05\%). This hints that the initial difference in performance is mainly attributed to difficulties in training (concatenated version has a much larger feature vector to manage).
    Note that evaluating the CTC shortcut branch yields poor decodings, despite the notable performance increase of the main network. For example, assuming line-level recognition and the padded/max-pooling setting, we report 5.26\% CER/19.76\% WER	for the validation set and 7.36\% CER/25.66\% WER for the test set.

    \item Applying all three modifications together achieves the best results across all setting and metrics.
    \item Word recognition reports improved results compared to line-level recognition with respect to the WER metric. This was expected, since word-level setting assumes perfect word segmentation. Interestingly enough, this is not the case for the CER metric. This can be explained by the lack of sufficient context (i.e. find a capital letter or a punctuation from the whole line information).
\end{itemize}

\begin{figure}[htb]
\centering
\includegraphics[width=.85\linewidth]{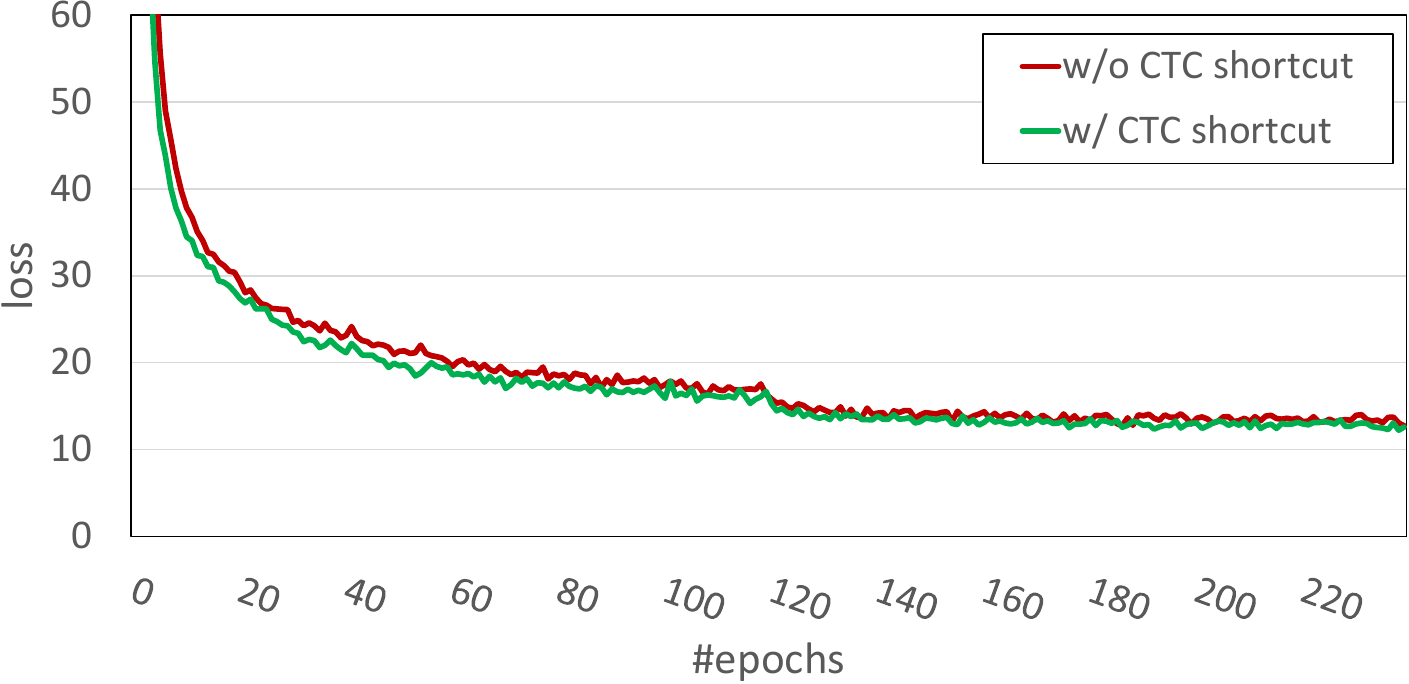} 
\caption{Behavior of HTR performance in terms of loss value with and without the extra CTC shortcut branch during the training phase. Reported curves correspond to the proposed line-level HTR system trained on the IAM dataset.} 
\label{fig:loss-comparison}
\end{figure}

\begin{figure}[htb]
\centering
\begin{tabular}{cc}
\includegraphics[width=.5\linewidth]{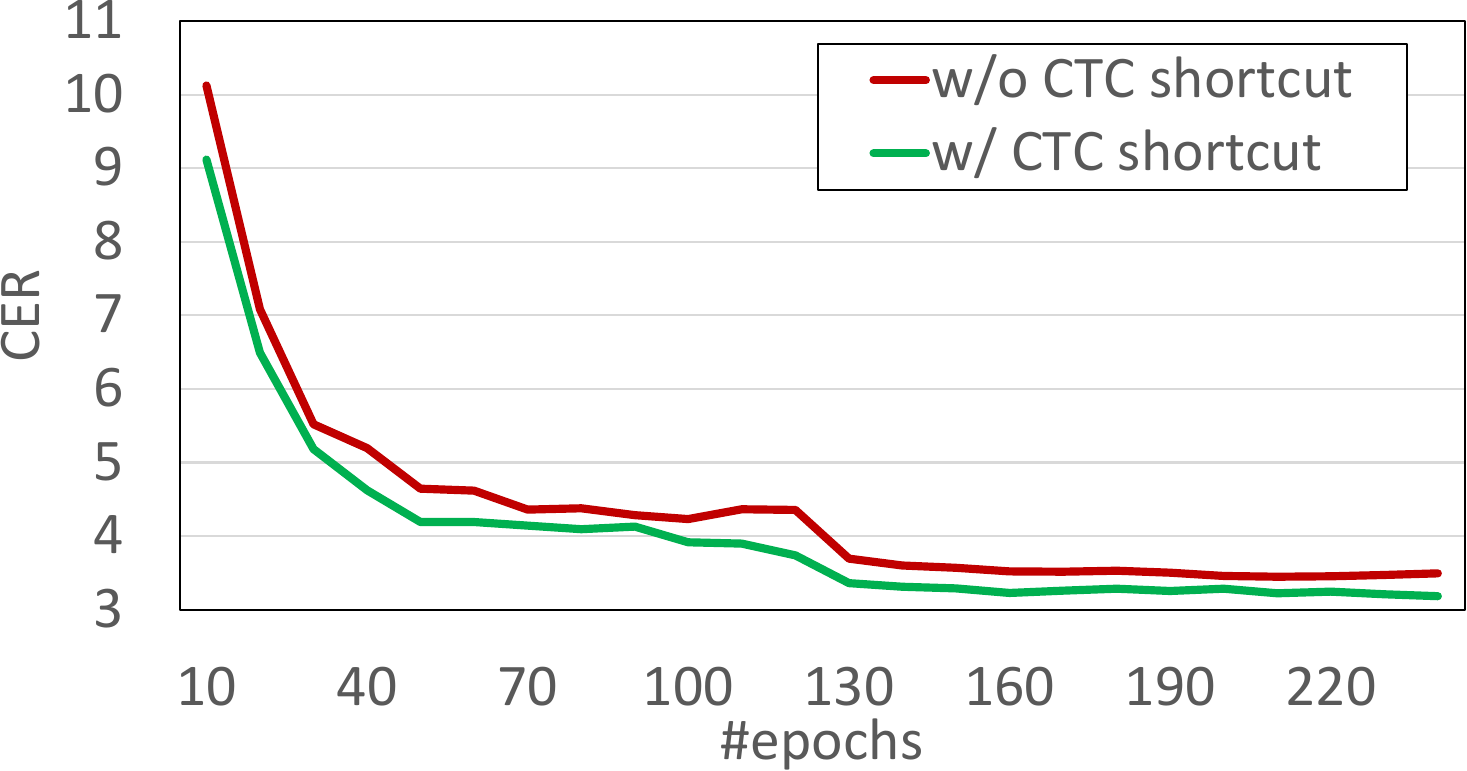} &
\includegraphics[width=.5\linewidth]{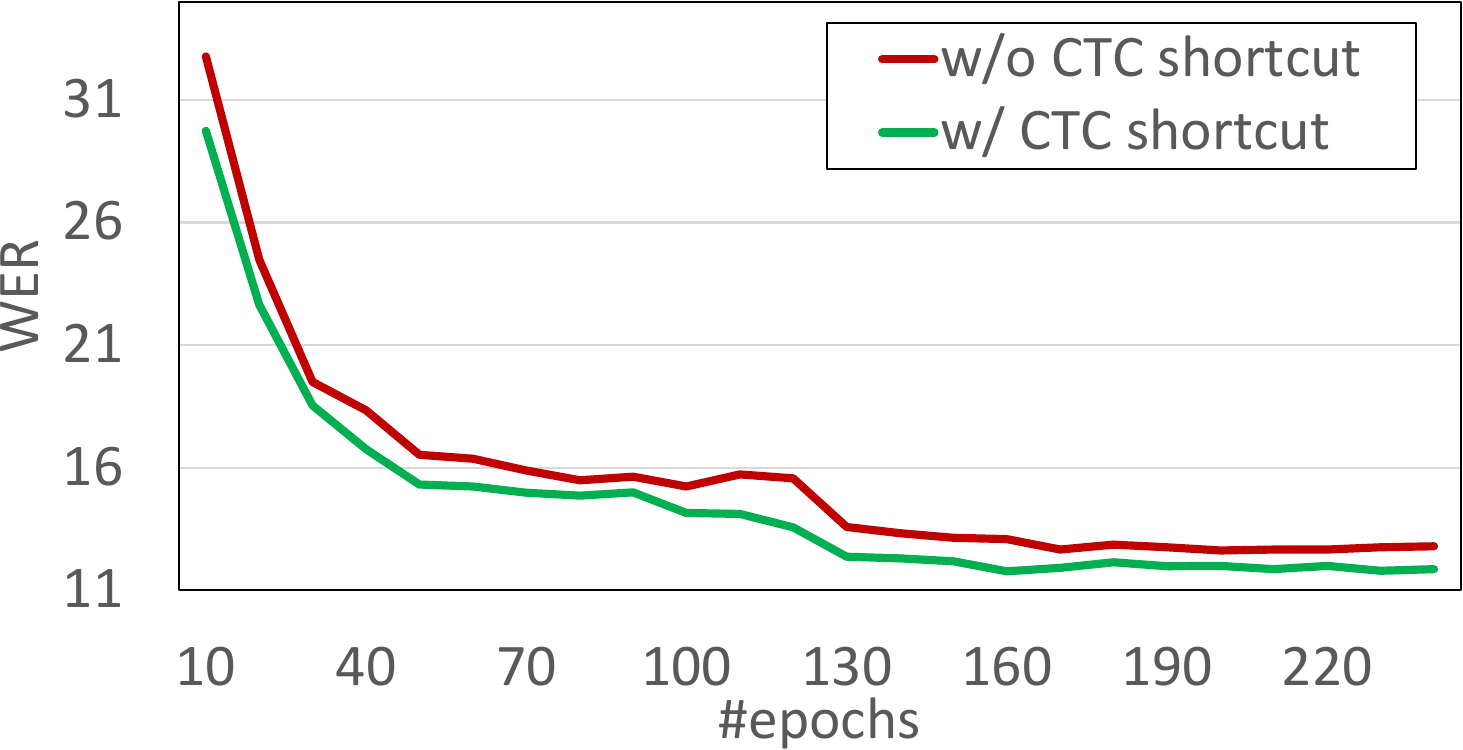} \\
\end{tabular}
\caption{Behavior of HTR performance in terms of CER (left) and WER (right) metrics with and without the extra CTC shortcut branch during the training phase. Reported curves correspond to the proposed line-level HTR system, trained on IAM dataset and evaluated on the validation set.} 
\label{fig:cwer-comparison}
\end{figure}

We also explore in more depth the CTC shortcut option, which seems to provide the best boost in performance. 
Specifically, we report the progress of both the loss and the CER/WER metrics (over the validation set) during the training procedure for the line-level recognition setting. 
The loss curves are depicted in Figure~\ref{fig:loss-comparison}, while the validation set evaluation metrics are reported in Figure~\ref{fig:cwer-comparison}. 
As we can see, loss curves are similar, but the case of CTC shortcut consistently has slightly better behavior.
The impact of the CTC shortcut is more clearly shown in CER/WER curves and thus solutions with greater generalization properties are expected when a model is trained along with the CTC shortcut.

\subsection{Comparison to State-of-the-Art Systems} 

Finally, we compare our method to several existing state-of-the-art methods, as shown in Table~\ref{table:soa_results}. The reported methods follow the same setting: line-level lexicon-free recognition. 
The proposed HTR system along with the suggested modifications achieves results comparable to the best performing methods. 
Notably, it outperforms the majority of existing works for both datasets and metrics despite many of the reported methods propose novel elements to further increase performance that are in general orthogonal to our approach. 
For example, the work of Chowdhury et al.~\cite{chowdhury2018efficient} presents better WER for the RIMES dataset while using a sequence-to-sequence approach (such models can produce increased WER as implicit language models can be learnt~\cite{retsinas2021seq2seq}), while our previous work~\cite{retsinas2021deformation} achieves better CER for the IAM dataset while using similar network (max-pooling flattening and padded input images) along with deformable convolutions and a post processing uncertainty reduction algorithm. 

Moreover, the very recent work of Luo et al.~\cite{luo2020learn} manages to outperform our method for the word-level recognition setting on the IAM dataset by using a STN component and a complex augmentation method, where "optimal" augmentations are learnt. Specifically, our method achieves 5.14\% CER / 14.33\%, while Luo et al. achieve  5.13\% CER / 13.35\% WER for the exact same setting. 
Nonetheless, their initial baseline network, stripped of all the extra modules (which could be added to the proposed architecture without any problem), performs poorly: 7.39\% CER and 19.12\% WER.

Overall, we achieve very competitive results (outperforming other existing lexicon-free methods for line-level recognition on IAM) by only using a typical convolutional-recurrent architecture along with a set of simple, yet intuitive and effective modifications, forming an effective set of best practice suggestions which can be applied to the majority of HTR systems. 

\begin{table}[htb]
\caption{Performance comparison for IAM/RIMES datasets (line-level recognition)
}
\centering
\begin{tabular}{C{.3\linewidth}C{.15\linewidth}C{.15\linewidth}|C{.15\linewidth}C{.15\linewidth}} 
\hline
& \multicolumn{2}{c}{IAM} & \multicolumn{2}{c}{RIMES}\\
\hline
Method &  CER(\%) & WER(\%) &  CER(\%) & WER(\%) \\
\hline 
Chen et al.~\cite{chen2017simultaneous} & 11.15 & 34.55 & 8.29 & 30.5\\
Pham et al.~\cite{pham2014dropout} & 10.8 & 35.1 & 6.8 & 28.5\\
Khrishnan et al.~\cite{krishnan2018word} & 9.78 & 32.89  & - & -\\
Chowdhury et al.~\cite{chowdhury2018efficient} & 8.10 & 16.70 & 3.59 & 9.60 \\
Puigcerver~\cite{puigcerver2017multidimensional} & 6.2 & 20.2 & 2.60 & 10.7 \\  
Khrishnan et al.~\cite{krishnan2018word} & 9.78 & 32.89  & - & -\\
Markou et al.~\cite{markou2020} & 6.14 & 20.04 & 3.34 & 11.23 \\
Dutta et al.~\cite{dutta2018improving} & 5.8 & 17.8 & 5.07 & 14.7 \\
Wick et al.~\cite{wick2021transformer} & 5.67 & - & - & - \\
Michael et al.~\cite{michael2019evaluating} & 5.24 & - & - & - \\
Tassopoulou et al.~\cite{tassopoulou2021enhancing} & 5.18 & 17.68 & - & -\\
Yousef et al.~\cite{yousef2020accurate} & 4.9 & - & - & -\\
Retsinas et al.~\cite{retsinas2021deformation} & 4.55 & 16.08 & 3.04 & 10.56\\
\hline
Proposed & 4.62	& 15.89 & 2.75 & 9.93\\
\hline 
\end{tabular}
\label{table:soa_results}
\end{table}

\section{Conclusions}

In this paper, we proposed a series of best practice modifications over typical convolutional-recurrent networks trained with CTC loss. Apart from presenting a fairly compact architecture based on residual blocks, we present three impactful modifications: 1) retain aspect-ratio of input images gathered in batches through a padding operation, 2) apply a column-wise max-pooling operation between the convolutional backbone and the recurrent head of a typical HTR architecture for reduced computational effort and increased performance and 3) enhance performance through a CTC shortcut during training in order to circumvent an end-to-end training over recurrent networks, which have been proven "difficult" to train in various settings.    
All proposed modifications have proven to be very helpful, 
considerably increasing the performance of the vanilla network. 
Overall, the proposed system achieves results in the ballpark of state-of-the-art, 
while being orthogonal to the majority of modern deep learning modules and approaches. 

\section*{Acknowledgments}

This research has been partially co - financed by the EU and Greek national funds through the Operational Program Competitiveness, 
Entrepreneurship and Innovation, under 
the calls : ``RESEARCH - CREATE - INNOVATE'', project \emph{Culdile} (code T1E$\Delta$K - 03785)
and ``OPEN INNOVATION IN CULTURE'', project \emph{Bessarion} (T6YB$\Pi$ - 00214).

\bibliographystyle{splncs04}
\bibliography{refs}

\begin{thebibliography}{10}
\providecommand{\url}[1]{\texttt{#1}}
\providecommand{\urlprefix}{URL }
\providecommand{\doi}[1]{https://doi.org/#1}

\bibitem{bishop2006pattern}
Bishop, C.M.: Pattern Recognition and Machine Learning. Springer (2006)

\bibitem{chen2017simultaneous}
Chen, Z., Wu, Y., Yin, F., Liu, C.L.: Simultaneous script identification and
  handwriting recognition via multi-task learning of recurrent neural networks.
  In: 14th IAPR International Conference on Document Analysis and Recognition
  (ICDAR). vol.~1, pp. 525--530. IEEE (2017)

\bibitem{chowdhury2018efficient}
Chowdhury, A., Vig, L.: An efficient end-to-end neural model for handwritten
  text recognition (2018)

\bibitem{collobert2019fully}
Collobert, R., Hannun, A., Synnaeve, G.: A fully differentiable beam search
  decoder. In: International Conference on Machine Learning. pp. 1341--1350.
  PMLR (2019)

\bibitem{dutta2018improving}
Dutta, K., Krishnan, P., Mathew, M., Jawahar, C.: Improving {CNN}-{RNN} hybrid
  networks for handwriting recognition. In: 2018 16th International Conference
  on Frontiers in Handwriting Recognition (ICFHR). pp. 80--85. IEEE (2018)

\bibitem{fischer2012lexicon}
Fischer, A., Keller, A., Frinken, V., Bunke, H.: Lexicon-free handwritten word
  spotting using character {HMM}s. Pattern Recognition Letters  \textbf{33}(7),
   934--942 (2012)

\bibitem{fischer2012handwriting}
Fischer, A.: Handwriting recognition in historical documents. Ph.D. thesis,
  Verlag nicht ermittelbar (2012)

\bibitem{graves2012connectionist}
Graves, A.: Connectionist temporal classification. In: Supervised Sequence
  Labelling with Recurrent Neural Networks, pp. 61--93. Springer (2012)

\bibitem{graves2006connectionist}
Graves, A., Fern{\'a}ndez, S., Gomez, F., Schmidhuber, J.: Connectionist
  temporal classification: labelling unsegmented sequence data with recurrent
  neural networks. In: Proceedings of the 23rd international conference on
  Machine learning. pp. 369--376 (2006)

\bibitem{greff2016lstm}
Greff, K., Srivastava, R.K., Koutn{\'\i}k, J., Steunebrink, B.R., Schmidhuber,
  J.: {LSTM}: A search space odyssey. IEEE Transactions on Neural Networks and
  Learning Systems  \textbf{28}(10),  2222--2232 (2016)

\bibitem{grosicki12008rimes}
Grosicki, E., Carre, M., Brodin, J.M., Geoffrois, E.: Rimes evaluation campaign
  for handwritten mail processing  (2008)

\bibitem{he2016deep}
He, K., Zhang, X., Ren, S., Sun, J.: Deep residual learning for image
  recognition. In: Proceedings of the IEEE Conference on Computer Vision and
  Pattern Recognition. pp. 770--778 (2016)

\bibitem{kingma2014adam}
Kingma, D.P., Ba, J.: Adam: A method for stochastic optimization. In:
  Proceedings of the International Conference on Learning Representations
  (ICLR) (2015)

\bibitem{krishnan2018word}
Krishnan, P., Dutta, K., Jawahar, C.: Word spotting and recognition using deep
  embedding. In: 2018 13th IAPR International Workshop on Document Analysis
  Systems (DAS). pp.~1--6. IEEE (2018)

\bibitem{leifert2016cells}
Leifert, G., Strau, T., Gr, T., Wustlich, W., Labahn, R., et~al.: Cells in
  multidimensional recurrent neural networks. Journal of Machine Learning
  Research  \textbf{17}(97),  1--37 (2016)

\bibitem{luo2020learn}
Luo, C., Zhu, Y., Jin, L., Wang, Y.: Learn to augment: Joint data augmentation
  and network optimization for text recognition. In: Proceedings of the
  IEEE/CVF Conference on Computer Vision and Pattern Recognition. pp.
  13746--13755 (2020)

\bibitem{markou2020}
Markou, K., Tsochatzidis, L., Zagoris, K., Papazoglou, A., Karagiannis, X.,
  Symeonidis, S., Pratikakis, I.: A convolutional recurrent neural network for
  the handwritten text recognition of historical greek manuscripts. In:
  International Workshop on Pattern Recognition for Cultural Heritage (PATRECH)
  (2020)

\bibitem{marti2002iam}
Marti, U.V., Bunke, H.: The iam-database: an english sentence database for
  offline handwriting recognition. International Journal on Document Analysis
  and Recognition  \textbf{5}(1),  39--46 (2002)

\bibitem{michael2019evaluating}
Michael, J., Labahn, R., Gr{\"u}ning, T., Z{\"o}llner, J.: Evaluating
  sequence-to-sequence models for handwritten text recognition. In: 2019
  International Conference on Document Analysis and Recognition (ICDAR). pp.
  1286--1293. IEEE (2019)

\bibitem{pham2014dropout}
Pham, V., Bluche, T., Kermorvant, C., Louradour, J.: Dropout improves recurrent
  neural networks for handwriting recognition. In: 2014 14th international
  conference on frontiers in handwriting recognition. pp. 285--290. IEEE (2014)

\bibitem{puigcerver2017multidimensional}
Puigcerver, J.: Are multidimensional recurrent layers really necessary for
  handwritten text recognition? In: 2017 14th IAPR International Conference on
  Document Analysis and Recognition (ICDAR). vol.~1, pp. 67--72. IEEE (2017)

\bibitem{retsinas2018transferable}
Retsinas, G., Sfikas, G., Gatos, B.: Transferable deep features for keyword
  spotting. In: Multidisciplinary Digital Publishing Institute Proceedings.
  vol.~2, p.~89 (2018)

\bibitem{retsinas2021iterative}
Retsinas, G., Sfikas, G., Nikou, C.: Iterative weighted transductive learning
  for handwriting recognition. In: International Conference on Document
  Analysis and Recognition. pp. 587--601. Springer (2021)

\bibitem{retsinas2021deformation}
Retsinas, G., Sfikas, G., Nikou, C., Maragos, P.: Deformation-invariant
  networks for handwritten text recognition. In: 2021 IEEE International
  Conference on Image Processing (ICIP). pp. 949--953. IEEE (2021)

\bibitem{retsinas2021seq2seq}
Retsinas, G., Sfikas, G., Nikou, C., Maragos, P.: From {Seq2Seq} recognition to
  handwritten word embeddings. In: Proceedings of the British Machine Vision
  Conference (BMVC) (2021)

\bibitem{sudholt2016phocnet}
Sudholt, S., Fink, G.A.: {PHOCNet}: A deep convolutional neural network for
  word spotting in handwritten documents. In: Proceedings of the {$15^{th}$}
  International Conference on Frontiers in Handwriting Recognition (ICFHR). pp.
  277--282 (2016)

\bibitem{sueiras2018offline}
Sueiras, J., Ruiz, V., Sanchez, A., Velez, J.F.: Offline continuous handwriting
  recognition using sequence to sequence neural networks. Neurocomputing
  \textbf{289},  119--128 (2018)

\bibitem{tassopoulou2021enhancing}
Tassopoulou, V., Retsinas, G., Maragos, P.: Enhancing handwritten text
  recognition with n-gram sequence decomposition and multitask learning. In:
  2020 25th International Conference on Pattern Recognition (ICPR). pp.
  10555--10560. IEEE (2021)

\bibitem{vaswani2017attention}
Vaswani, A., Shazeer, N., Parmar, N., Uszkoreit, J., Jones, L., Gomez, A.N.,
  Kaiser, {\L}., Polosukhin, I.: Attention is all you need. In: Advances in
  neural information processing systems. pp. 5998--6008 (2017)

\bibitem{wick2021transformer}
Wick, C., Z{\"o}llner, J., Gr{\"u}ning, T.: Transformer for handwritten text
  recognition using bidirectional post-decoding. In: International Conference
  on Document Analysis and Recognition. pp. 112--126. Springer (2021)

\bibitem{wigington2017data}
Wigington, C., Stewart, S., Davis, B., Barrett, B., Price, B., Cohen, S.: Data
  augmentation for recognition of handwritten words and lines using a cnn-lstm
  network. In: 2017 14th IAPR International Conference on Document Analysis and
  Recognition (ICDAR). vol.~1, pp. 639--645. IEEE (2017)

\bibitem{yousef2020accurate}
Yousef, M., Hussain, K.F., Mohammed, U.S.: Accurate, data-efficient,
  unconstrained text recognition with convolutional neural networks. Pattern
  Recognition  \textbf{108},  107482 (2020)

\end{thebibliography}


\begin{thebibliography}{1}

\bibitem{kour2014real}
George Kour and Raid Saabne.
\newblock Real-time segmentation of on-line handwritten arabic script.
\newblock In {\em Frontiers in Handwriting Recognition (ICFHR), 2014 14th
  International Conference on}, pages 417--422. IEEE, 2014.

\bibitem{kour2014fast}
George Kour and Raid Saabne.
\newblock Fast classification of handwritten on-line arabic characters.
\newblock In {\em Soft Computing and Pattern Recognition (SoCPaR), 2014 6th
  International Conference of}, pages 312--318. IEEE, 2014.

\bibitem{hadash2018estimate}
Guy Hadash, Einat Kermany, Boaz Carmeli, Ofer Lavi, George Kour, and Alon
  Jacovi.
\newblock Estimate and replace: A novel approach to integrating deep neural
  networks with existing applications.
\newblock {\em arXiv preprint arXiv:1804.09028}, 2018.

\end{thebibliography}

\end{document}